\documentclass[oribibl]{llncs}
\usepackage{llncsdoc}

\usepackage{epsfig}
\usepackage{amssymb}
\usepackage{amstext}
\usepackage{amsmath}

\input xy
\xyoption{all}

\usepackage{algorithm}
\usepackage{algorithmic}

\usepackage{subfigure}
\DeclareGraphicsExtensions{.eps}

 \newtheorem{prop}{Proposition}

\begin{document}

\title{\uppercase{Symbolic Knowledge Extraction using {\L}ukasiewicz Logics}}
\author{Carlos Leandro}
\institute{Departamento de Matem\'{a}tica,\\
Instituto Superior de Engenharia de Lisboa, Portugal.\\
\email{miguel.melro.leandro@gmail.com}}
%
%
\maketitle

\begin{abstract}
This work describes a methodology that combines logic-based systems and connectionist systems. Our approach uses finite truth-valued {\L}ukasiewicz logic, wherein every connective can be defined by a neuron in an artificial network \cite{Castro98}. This allowed the injection of first-order formulas into a network architecture, and also simplified symbolic rule extraction. For that we trained a neural networks using the Levenderg-Marquardt algorithm, where we restricted the knowledge dissemination in the network structure. This procedure reduces neural network plasticity without drastically damaging the learning performance, thus making the descriptive power of produced neural networks similar to the descriptive power of {\L}ukasiewicz logic language and simplifying the translation between symbolic and connectionist structures. We used this method for reverse engineering truth table and in extraction of formulas from real data sets.
\end{abstract}
 ----------------------------------------------------------------------
\section{\uppercase{Introduction}}

There are essentially two representation paradigms, usually taken very differently. On one hand, symbolic-based descriptions are specified through a grammar that has fairly clear semantics. On the other hand, the usual way to see information presented using a connectionist description is its codification on a neural network (NN). Artificial NNs, in principle, combine - among other things - the ability to learn and robustness or insensitivity to perturbations of input data.
NNs are usually taken as black boxes, thereby providing little insight into how the information is codified. It is natural to seek a synergy integrating the \emph{white-box} character of symbolic base representation and the learning power of artificial neuronal networks. Such neuro-symbolic models are currently a very active area of research: for the extraction of logic programs from trained networks see \cite{Hitzler04} \cite{Holldobler00}.

Our approach to neuro-symbolic models and knowledge extraction is based on a comprehensive language for humans, representable directly in a NN topology and able to be used. This is done on knowledge-based networks \cite{Fu93}  \cite{Towell94}, to generate the initial network architecture from crude symbolic domain knowledge. In the other direction, the hardest problem, neural language can be translated into a symbolic language. However in \cite{Gallan88} \cite{Gallan94} \cite{Towell93} this processes is used by identifing the most significant determinants of decision or classification. Hence, any individual unit must be associated with a single concept or feature of the problem domain. In this work we used a first-order language wherein formulas are interpreted as NNs. In this framework formulas are simple to inject into a multilayer feed-forward network, and the system is free from the need of giving interpretation to hidden units in the problem domain.

 Our approx to the generation of neuro-symbolic models used {\L}ukasiewicz logic. This type of many-valued logic has a very useful property motivated by the ''linearity'' of logic connectives. Every logic connective can be defined by a neuron in an artificial network having, by activation function, the identity truncated to zero and one \cite{Castro98}. This allows the direct codification of formulas into network architecture, and simplifies the extraction of rules. Multilayer feed-forward NN, having this type of activation function,  can be trained efficiently using the Levenderg-Marquardt (LM) algorithm \cite{HaganMenhaj99}, and the generated network can be simplified quickly using the "Optimal Brain Surgeon" algorithm proposed by B. Hassibi, D. G. Stork and G.J. Stork \cite{Hassibi93}.

This strategy has good performance when applied to the reconstruction of formulas from truth tables. In this type of reverse engineering problem, we presuppose no noise. However, the process is stable for the introduction of Gaussian noise. This motivates its application to extract comprehensible symbolic rules from real data.

\section{\uppercase{Preliminaries}}

\subsection{{\L}ukasiewicz logics}
Classical propositional logic is one of the earliest formal systems of logic. The algebraic semantics of this logic are given by Boolean algebra. Both, the logic and the algebraic semantics have been generalized in many directions. Many-valued logics, is one of this generalizations, and can be conceived as a set of formal representation languages that proven to be useful for both real world and computer science applications. In applications of many-valued logic, like fuzzy logic, the properties of Boolean conjunction are too rigid, this is overtake extending a new binary connective, $\otimes$, usually called \emph{fusion}. The generalization of Boolean algebra can be based in the relationship between conjunction and implication given by
\[
_{(x\otimes y)\leq z \Leftrightarrow x\leq (y \Rightarrow z) \Leftrightarrow y\leq (x \Rightarrow z).}
\]
These equivalences, can be used to present implication as a generalized inverse for conjunction.

These two operators are defined in a partially ordered set of truth values, $(P,\leq)$, thereby extending the two-valued set of an Boolean algebra. If $P$ has more than two values, the associated logics are called a \emph{many-valued logics}. A many-valued logic having $[0,1]$ as set of truth values is called a \emph{fuzzy logic}. In this type of logics a continuous fusion operator $\otimes$ is known as a \emph{t}-norm. The following are example of continuous $t$-norms:
\begin{enumerate}
  \item \emph{{\L}ukasiewicz} $t$-norm: $x\otimes y=\max(0,x+y-1)$.
  \item Product $t$-norm: $x\otimes y=xy$ usual product between real numbers.
  \item G\"{o}del $t$-norm: $x\otimes y=\min(x,y)$.
\end{enumerate}
The fuzzy logic defined using \emph{{\L}ukasiewicz} $t$-norm is called {\L}ukasiewicz logic ({\L}logic) and the corresponding propositional calculus has a nice complete axiomatization \cite{Hajek95}. In this type of logic the implication, is called \emph{residuum} operator, and is given by $x\Rightarrow y=\min(1,1-x+y)$.

Like first-order languages, in {\L}logic, sentences are usually built from (countable) set of propositional variables,   $\otimes$ the fusion operator,  implication $\Rightarrow$, and the truth constant 0.  Further connectives are defined as follows:
\begin{center}
\begin{tabular}{lcl}
  $\neg\varphi_1:= \varphi_1\Rightarrow 0,$ & &$1:=0\Rightarrow 0$ \\
  $\varphi_1\oplus\varphi_2:=\neg\varphi_1\Rightarrow\varphi_2,$ & & $\varphi_1\Leftrightarrow\varphi_2 := (\varphi_1\Rightarrow\varphi_2)\otimes(\varphi_2\Rightarrow\varphi_1)$ \\
\end{tabular}
\end{center}
The interpretation for a well-formed formula $\varphi$ is defined as usual, by assigning a truth value to each propositional variable.

\subsection{Processing units}

As mentioned in \cite{Amato02} there is a lack of a deep investigation of the relationships between logics and NNs. In this work we present a methodology using NNs to learn formulas from data.

In \cite{Castro98} it is shown how, by taking as activation function, $\psi$, the identity truncated to zero and one,
\[
_{\psi(x)=\min(1,\max(x, 0)),}
\]
it is possible to represent the corresponding NN as a combination of propositions of {\L}ukasiewicz calculus and \emph{viceversa} \cite{Amato02}.

\begin{figure}[h]
\begin{center}
\tiny
$
\xymatrix @R=5pt @C=5pt { x\ar@{-}[dr]_{w_1} & \ar@{-}[d]^{b} & \\
                           & *+[o][F-]{\psi} \ar@{-}[r]&z\;\;\Leftrightarrow\;\; z=\min(1,\max(0,w_1x+w_2y+b)) \\
           y\ar@{-}[ur]^{w_2} &  & \;\;=\psi_b(w_1x,w_2y) \
           }
$
\end{center}
\caption{functional interpretation for a NN }\label{interpretation}
\end{figure}
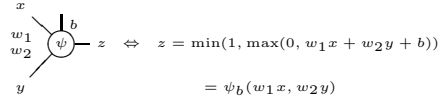

However, if we want apply NNs to learn {\L}ukasiewicz sentences, it seems more promising the use of a non-recursive approach to proposition evaluation. We can do this by defining the first-order language  as a set of circuits generated from the plugging of atomic components. For this, we used the library of components presented in table \ref{semioticaLuk}, interpreted as neural units and linked them together, to form NNs having only one output, without loops. These NNs are interpretation for formulas, having its structure where each neuron defines the connective identified by its label. This task of construct complex structures based on simplest ones can be formalized using generalized programming \cite{Fiadeiro97}.
\begin{table}
\begin{center}
\tiny
\begin{tabular}{|c|c||c|c||c|c||c|c|}
  \hline
  Formula: & Configuration: & Formula: & Configuration: & Formula: & Configuration: & Formula: & Configuration: \\
  \hline
  \hline
  $\neg x\oplus y$
  &
  $\xymatrix @R=6pt @C=6pt { x\ar@{-}[dr]_{-1} & \ar@{-}[d]^{1} & \\
                           & *+[o][F-]{\varphi} \ar@{-}[r]& \\
           y\ar@{-}[ur]^{1} &  &  \
           }$
  &
  $ x\otimes \neg y$
  &
  $\xymatrix @R=6pt @C=6pt { x\ar@{-}[dr]_{1} & \ar@{-}[d]^{0} & \\
                           & *+[o][F-]{\varphi} \ar@{-}[r]& \\
           y\ar@{-}[ur]^{-1} &  &  \
           }$
  &
  $x\oplus y$
  &
  $\xymatrix @R=6pt @C=6pt { x\ar@{-}[dr]_{1} & \ar@{-}[d]^{0} & \\
                           & *+[o][F-]{\varphi} \ar@{-}[r]& \\
           y\ar@{-}[ur]^{1} &  &  \
           }$
  &
  $\neg x\otimes \neg y)$
  &
  $\xymatrix @R=6pt @C=6pt { x\ar@{-}[dr]_{-1} & \ar@{-}[d]^{1} & \\
                          & *+[o][F-]{\varphi} \ar@{-}[r]& \\
           y\ar@{-}[ur]^{-1} &  &  \
           }$ \\
  \hline
  $x\oplus \neg y$ & $\xymatrix @R=6pt @C=6pt { x\ar@{-}[dr]_{1} & \ar@{-}[d]^{1} & \\
                           & *+[o][F-]{\varphi} \ar@{-}[r]& \\
           y\ar@{-}[ur]^{-1} &  &  \
           }$ & $x\otimes y$ & $\xymatrix @R=6pt @C=6pt { x\ar@{-}[dr]_{1} & \ar@{-}[d]^{-1} & \\
                           & *+[o][F-]{\varphi} \ar@{-}[r]& \\
           y\ar@{-}[ur]^{1} &  &  \
           }$ &
  $\neg x\otimes y$ & $\xymatrix @R=6pt @C=6pt { x\ar@{-}[dr]_{-1} & \ar@{-}[d]^{0} & \\
                           & *+[o][F-]{\varphi} \ar@{-}[r]& \\
           y\ar@{-}[ur]^{1} &  &  \
           }$ &  &  \\
           \hline
\end{tabular}
\end{center}
\caption{Possible configurations for a neuron in a {\L}NN a its interpretation.}\label{semioticaLuk}
\end{table}
The neurons of these types of networks, which have two inputs and one output, can be interpreted as a function (see figure \ref{interpretation}) and are generically denoted, in the following, by $\psi_b(w_1x_1,w_2x_2)$, where $b$ represent the bias, $w_1$ and $w_3$ are the weights and, $x_1$ and $x_2$ input values. In this context a network is the \emph{functional interpretation} of a sentence in the string-based notation when the relation, defined by network execution, corresponds to the sentence truth table.
The use of NNs as interpretation of formulas simplifies the transformation between string-based representations and the network representation, allowing one to write:

\begin{prop}\label{prop1}
Every well-formed formula in the {\L}logic language can be codified using a NN, and the network defines the formula interpretation, when the activation function is the identity truncated to zero and one.
\end{prop}

For instance, the semantic for sentence $_{\varphi=(x\otimes y\Rightarrow z)\oplus(z \Rightarrow w),}$ can be described using the bellow network or can be codified by the presented set of matrices. From this matrices we must note that the partial interpretation of each unit can be seen as a simple exercise of pattern checking, where we must take by reference relation, between formulas and configuration, described in table \ref{semioticaLuk}.
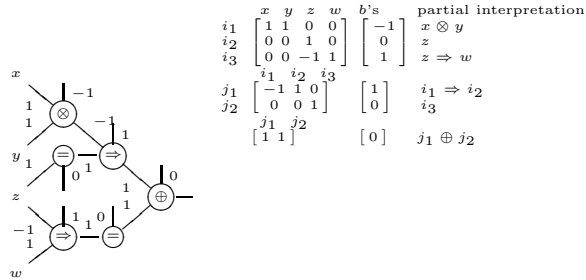
\begin{center}
\tiny
$
\xymatrix @R=6pt @C=6pt { x\ar@{-}[dr]_{1} & \ar@{-}[d]^{-1} & \\
                           & *+[o][F-]{\otimes} \ar@{-}[rd]^{-1}& \ar@{-}[d]^{1} \\
           y\ar@{-}[ur]^{1} & *+[o][F-]{=}\ar@{-}[r]_{1}  &  *+[o][F-]{\Rightarrow} \ar@{-}[rd]_{1} &\ar@{-}[d]^{0}\\
           z\ar@{-}[dr]_{-1} \ar@{-}[ur]^{1}& \ar@{-}[d]^{1}\ar@{-}[u]_{0} & \ar@{-}[d]_{0} &*+[o][F-]{\oplus} \ar@{-}[r]&\\
                           & *+[o][F-]{\Rightarrow} \ar@{-}[r]^{1}&*+[o][F-]{=} \ar@{-}[ru]^{1}\\
           w\ar@{-}[ur]^{1} &  &  \\
           }
$
 \begin{tabular}{llll}
 &$\begin{array}{cccc}
      \;x &\; y &\; z &\; w \\
    \end{array}$ & $b$'s & \tiny partial interpretation
      \\
$ \begin{array}{c}
      i_1  \\
      i_2 \\
      i_3 \\
    \end{array}$
    &
  $\left[
    \begin{array}{cccc}
      1 &  1 & 0 & 0 \\
      0 &  0 & 1 & 0 \\
      0 &  0 & -1 & 1\\
    \end{array}
  \right]
  $ &$ \left[
       \begin{array}{c}
         -1 \\
         0 \\
         1 \\
       \end{array}
     \right]
    $& $\begin{array}{l}
         x \otimes y \\
         z  \\
         z\Rightarrow w\\
       \end{array}$\\
    &$\begin{array}{ccc}
      \;i_1 & \;i_2 & \;i_3 \\
    \end{array}$& &
       \\
       $
  \begin{array}{c}
      j_1 \\
      j_2 \\
    \end{array}$
  &
  $\left[
    \begin{array}{ccc}
      -1 &  1 & 0 \\
       0 &  0 & 1 \\
    \end{array}
  \right]
   $&$ \left[
       \begin{array}{c}
         1 \\
         0 \\
       \end{array}
     \right]
    $& $\begin{array}{l}
         i_1\Rightarrow i_2 \\
        i_3 \\
       \end{array}$\\
   &$\begin{array}{cc}
         \;j_1 &\;j_2 \\
    \end{array}$
       \\
  &$\left[
      \begin{array}{cc}
         1 & 1 \\
       \end{array}
     \right] $& $\left[
       \begin{array}{c}
        0 \\
       \end{array}
     \right]$ & $j_1\oplus j_2$\\
\end{tabular}
\end{center}
\begin{center}
\tiny
\begin{tabular}{lll}
   INTERPRETATION:&\\
   &$j_1\oplus j_2=(i_1\Rightarrow i_2)\oplus (i_3)=((x \otimes y)\Rightarrow z)\oplus (z\Rightarrow w)$\\
\end{tabular}
\end{center}
In this sense this NN can be seen as an interpretation for sentence $\varphi$; it codifies $f_\varphi$, the proposition truth table.
\[
_{f_\varphi(x,y,z,w)=\psi_0(\psi_0(\psi_1(-z,w)),\psi_1(\psi_0(z),-\psi_{-1}(x,y)))}
\]
However truth table $f_\varphi$ is a continuous structure, for our goal, it must be discretized using a finite structure, ensuring sufficient information to describe the original formula. A truth table $f_\varphi$ for a formula $\varphi$, in a fuzzy logic, is a map $f_\varphi:[0,1]^m\rightarrow [0,1]$, where $m$ is the number of propositional variables used in $\varphi$. For each integer $n>0$, let $S_n$ be the set $\{0,\frac{1}{n},\ldots,\frac{n-1}{n},1\}$. Each $n>0$, defines a sub-table for $f_\varphi$ defined by $f_\varphi^{(n)}:(S_n)^m\rightarrow [0,1]$, given by $f_\varphi^{(n)}(\bar{v})=f_\varphi(\bar{v})$, and called the $\varphi$ \emph{(n+1)-valued truth sub-table}.

\subsection{Similarity between a configuration and a formula}

We call a \emph{Castro neural network} (CNN) a type of NN having as activation function $\psi(x)=\min(1,max(0,x))$, where its weights are -1, 0 or 1 and having by bias an integer. A CNN is called \emph{{\L}ukasiewicz neural network} ({\L}NN) if it can be codified as a binary NN: i.e. a  CNN where each neuron has one or two inputs. A network is called \emph{un-representable} if is impossible to codify using a binary CNN. Note that, a binary CNN  can be translated directly into {\L}ukasiewicz firs-order language, using the correspondences described in table \ref{semioticaLuk}.

Below we present functional interpretation for formulas defined using a neuron with two inputs. These interpretation are classified as disjunctive interpretations ou conjunctive interpretations.
\begin{center}
\tiny
\begin{tabular}{|c|c|}
  \hline
  & \\
  Disjunctive interpretations & Conjunctive interpretations \\
  & \\
  \hline
  $\psi_0(x_1,x_2)=f_{x_1\oplus x_2}$, $\;\;\psi_1(x_1,-x_2)=f_{x_1\oplus \neg x_2}$& $\psi_{-1}(x_1,x_2)=f_{x_1\otimes x_2}$, $\;\;\psi_0(x_1,-x_2)=f_{x_1\otimes \neg x_2}$\\
  $\psi_1(-x_1,x_2)=f_{\neg x_1\oplus  x_2}$, $\;\;\psi_2(-x_1,-x_2)=f_{\neg x_1\oplus \neg x_2}$ & $\psi_0(-x_1,x_2)=f_{\neg x_1\otimes x_2}$, $\;\;\psi_1(-x_1,-x_2)=f_{\neg x_1\otimes \neg x_2}$\\
  \hline
\end{tabular}
\end{center}
These correspond to all possible configurations of neurons with two inputs. The other possible configurations are constant and can also be seen as representable configurations. For instance, $_{\psi_{b}(x_1,x_2)=0}$, if $b<-1$, and
$_{\psi_{b}(-x_1,-x_2)=1}$, if $b>1$.

In this sense, every representable network can be codified by a NN where the neural units satisfy one of the above patterns. Below we can see also examples of representable configurations for a neuron with three inputs.  In the table we presente how they can be codified using representable NNs having units with two inputs, and the corresponding interpreting formula in the sting-based notation.
\begin{center}
\tiny
\begin{tabular}{|c|}
  \hline
   \\
   Conjunctive configurations \\
   \\
  \hline
  $\psi_{-2}(x_1,x_2,x_3)=\psi_{-1}(x_1,\psi_{-1}(x_2,x_3))=f_{x_1\otimes x_2\otimes x_3}$ \\
  $\psi_{-1}(x_1,x_2,-x_3)=\psi_{-1}(x_1,\psi_{0}(x_2,-x_3))=f_{x_1\otimes x_2\otimes \neg x_3}$\\
  $\psi_{0}(x_1,-x_2,-x_3)=\psi_{-1}(x_1,\psi_{1}(-x_2,-x_3))=f_{x_1\otimes \neg x_2\otimes \neg x_3}$ \\
  $\psi_{1}(-x_1,-x_2,-x_3)=\psi_{0}(-x_1,\psi_{1}(-x_2,-x_3))=f_{\neg x_1\otimes \neg x_2\otimes \neg x_3}$ \\
  \hline
\end{tabular}
\end{center}
\begin{center}
\tiny
\begin{tabular}{|c|}
  \hline
   \\
 Disjunctive interpretations \\
   \\
  \hline
  $\psi_{0}(x_1,x_2,x_3)=\psi_{0}(x_1,\psi_{0}(x_2,x_3))=f_{x_1\oplus x_2\oplus x_3}$ \\
  $\psi_{1}(x_1,x_2,-x_3)=\psi_{0}(x_1,\psi_{1}(x_2,-x_3))=f_{x_1\oplus x_2\oplus \neg x_3}$ \\
  $\psi_{2}(x_1,-x_2,-x_3)=\psi_{0}(x_1,\psi_{2}(-x_2,-x_3))=f_{x_1\oplus \neg x_2\oplus \neg x_3}$ \\
  $\psi_{3}(-x_1,-x_2,-x_3)=\psi_{1}(-x_1,\psi_{2}(-x_2,-x_3))=f_{\neg x_1\oplus \neg x_2\oplus \neg x_3}$ \\
  \hline
\end{tabular}
\end{center}
Constant configurations like $_{\psi_{b}(x_1,x_2,x_3)=0}$, if $b<-2$, and $_{\psi_{b}(-x_1,-x_2,-x_3)=1}$, if $b>3$, are also representable. However there are examples of un-representable networks with three inputs like the configuration $_{\psi_{b}(-x_1,x_2,x_3)}$.

Naturally, a neuron configuration - when representable - can by codified by
different structures using a {\L}NN. Particularly, we have:
\begin{prop}
If the neuron configuration $_{\alpha=\psi_b(x_1,x_2,\ldots,x_{n-1},x_n)}$ is representable, but not constant, it can be codified in a {\L}NN  with the following structure:\\
$
_{\beta=\psi_{b_1}(x_1,\psi_{b_2}(x_2,\ldots,\psi_{b_{n-1}}(x_{n-1},x_n)\ldots)).}
$
\end{prop}

And, since the $n$-nary operator $\psi_b$ is commutative, variables could interchange its position in function $\beta$ without changing the operator output. By this we mean that, in the string-based representation, variable permutation generates equivalent formulas. From this we concluded:

\begin{prop}
If $_{\alpha=\psi_b(x_1,x_2,\ldots,x_{n-1},x_n)}$ is representable, but not constant, it is the interpretation of a disjunctive formula or a conjunctive formula.
\end{prop}

Recall that disjunctive formulas are written using only disjunctions and negations, and conjunctive formulas are written using only conjunctions and negations. This leave us with the task of classifying a neuron configuration according to its representation. For that, we established a relationship using the configuration bias and the number of negative and positive weights.
\begin{prop}\label{conf classification}
Given the neuron configuration
$
_{\alpha=\psi_b(-x_1,-x_2,\ldots,-x_n, x_{n+1},\ldots,x_m)}
$
with $m=n+p$ inputs and where $n$ and $p$ are, respectively, the number of negative and the number of positive weights, on the neuron configuration:
\begin{enumerate}
  \item If $b=-(m-1)+n$ (i.e. $b=-p+1$) the neuron is called a \emph{conjunction} and it is a interpretation for
$
_{\neg x_1\otimes\ldots\otimes\neg x_n\otimes x_{n+1}\otimes\ldots\otimes x_m.}
$
  \item When $b=n$ the neuron is called a \emph{disjunction} and it is a interpretation of
$
_{\neg x_1\oplus\ldots\oplus\neg x_n\oplus x_{n+1}\oplus\ldots\oplus x_m.}
$
\end{enumerate}
\end{prop}

From the structure associated with this type of formula, we proposed the following structural characterization for representable neurons:

\begin{prop}
Every conjunctive or disjunctive configuration \\ $\alpha=\psi_b(x_1,x_2,\ldots,x_{n-1},x_n)$, can be codified by a  {\L}NN \\ $_{\beta=\psi_{b_1}(x_1,\psi_{b_2}(x_2,\ldots,\psi_{b_{n-1}}(x_{n-1},x_n)\ldots)),}$
where $_{b=b_1+b_2+\cdots+b_{n-1}\text{ and }b_1\leq b_2\leq \cdots\leq b_{n-1}.}$
\end{prop}

This property can be translated in the following neuron rewriting rule,
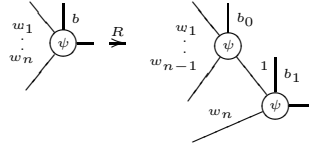
\[
\tiny
\xymatrix @R=6pt @C=6pt {  \ar@{-}[rd]_{w_1} & \ar@{-}[d]^{b} &    &  \\
                      \vdots     & *+[o][F-]{\psi} \ar@{-}[r] & \ar[r]^{R} &\\
            \ar@{-}[ru]^{w_n} & &\\
           }
\xymatrix @R=7pt @C=7pt {  \ar@{-}[rd]_{w_1} & \ar@{-}[d]^{b_0} &    &  \\
                   \vdots        & *+[o][F-]{\psi} \ar@{-}[rd]^{1} & \ar@{-}[d]^{b_1}\\
            \ar@{-}[ru]^{w_{n-1}} & &*+[o][F-]{\psi}\ar@{-}[r]&\\
            \ar@{-}[rru]^{w_{n}} & &\\
           }
\]
linking equivalent networks, when values $b_0$ and $b_1$ satisfy $b=b_0+b_1$ and $b_1\leq b_0$, and are such that neither of the involved neurons have constant output.
This rewriting rule can be used to join equivalent configurations like:
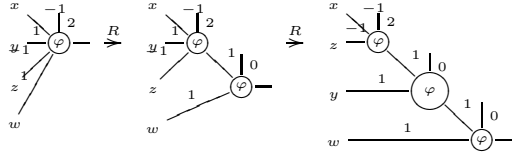
\[
\tiny
\xymatrix @R=6pt @C=6pt { x\ar@{-}[dr]^{-1} & \ar@{-}[d]^{2}                & \\
           y\ar@{-}[r]^{1} & *+[o][ F-]{\varphi} \ar@{-}[r] & \ar[r]^{R}&\\
           z\ar@{-}[ur]^{-1} &  & \\
           w\ar@{-}[uur]^{1} &  &  \\
           }
\xymatrix @R=6pt @C=6pt { x\ar@{-}[dr]^{-1} & \ar@{-}[d]^{2}                & \\
           y\ar@{-}[r]^{1} & *+[o][ F-]{\varphi} \ar@{-}[dr]^{1} & \ar@{-}[d]^{0} &\ar[r]^{R}&\\
           z\ar@{-}[ur]^{-1} &  & *+[o][ F-]{\varphi} \ar@{-}[r]&\\
           w\ar@{-}[urr]^{1} &  &  &\\
           }
\xymatrix @R=6pt @C=6pt { x\ar@{-}[dr]^{-1} & \ar@{-}[d]^{2}                  &                                &                                 &\\
                            z\ar@{-}[r]^{-1}  & *+[o][ F-]{\varphi} \ar@{-}[dr]^{1} & \ar@{-}[d]^{0}                 &                                  &\\
                            y\ar@{-}[rr]^{1}  &                                 & *++[o][ F-]{\varphi} \ar@{-}[dr]^{1}& \ar@{-}[d]^{0}                   &\\
                            w\ar@{-}[rrr]^{1} &                                 &                                & *+[o][ F-]{\varphi} \ar@{-}[r]  &\\
           }
\]
Note that, a representable CNN can be transformed by the application of rule R in a set of equivalent {\L}NN with simplest neuron configuration:
\begin{prop}
Un-representable neuron configurations are those transformed by
rule R in, at least, two non-equivalent NNs.
\end{prop}

For instance, the un-representable configuration $\psi_0(-x_1,x_2,x_3)$, is transformed by rule R in three non-equivalent configurations:
\begin{center}
\tiny
\begin{tabular}{|c|c|}
  \hline
  & \\
  $\psi_0(x_3,\psi_0(-x_1,x_2))=f_{x_3\oplus(\neg x_1\otimes x_2)}$ & $\psi_{-1}(x_3,\psi_{1}(-x,x_2))=f_{x_3\otimes(\neg x_1\otimes x_2)}$ \\ $\psi_0(-x_1,\psi_0(x_2,x_3))=f_{\neg x_1\otimes(x_2\oplus x_3)}$ &\\
  & \\
  \hline
\end{tabular}
\end{center}
The representable configuration $\psi_2(-x_1,-x_2,x_3)$ is transformed by rule R on only two distinct but equivalent configurations:
\begin{center}
\tiny
\begin{tabular}{|c|c|}
  \hline
  & \\
  $\psi_0(x_3,\psi_2(-x_1,-x_2))=f_{x_3\oplus \neg (x_1\otimes x_2)}$ & $\psi_1(-x_2,\psi_1(-x_1,x_3))=f_{\neg x_2\oplus (\neg x_1\oplus x_3)}$ \\
  & \\
  \hline
\end{tabular}
\end{center}

For the extraction of knowledge from trained NNs, we  translate neuron configuration in propositional connectives to form formulas. However, not all neuron configurations can be translated in formulas, but they can be approximate by formulas. To quantify the approximation quality we defined the notion of interpretation $\lambda$-similar to a formula.

Two neuron configurations $\alpha=\psi_{b}(x_1,x_2,\ldots,x_n)$ and $\beta=\psi_{b'}(y_1,y_2,\ldots,y_n)$, are  called $\lambda$-similar,  in a $(m+1)$-valued {\L}logic, if $\lambda$ is the exponential of mean absolute error symmetric, evaluated taking the same cases in the truth sub-table of $\alpha$ and  $\beta$. When we have $_{\lambda=e^{-\sum_{\bar{x}\in T}\frac{|\alpha(\bar{x})-\beta(\bar{x})|}{\sharp T},}}$ write
$_{\alpha\sim_\lambda\beta.}$

If $\alpha$ is un-representable and $\beta$ is representable, the second configuration is called \emph{a representable approximation} to the first.

On the $2$-valued {\L}logic (the Boolean logic case), we have for the un-representable configuration $\alpha=\psi_0(-x_1,x_2,x_3)$:
\begin{center}
\tiny
\begin{tabular}{|c|c|}
  \hline
  & \\
  $\psi_0(-x_1,x_2,x_3)\sim_{0.883}\psi_0(x_3,\psi_0(-x_1,x_2))$& $\psi_0(-x_1,x_2,x_3)\sim_{0.883}\psi_{-1}(x_3,\psi_{1}(-x_1,x_2))$\\
  $\psi_0(-x_1,x_2,x_3)\sim_{0.883}\psi_0(-x_1,\psi_0(x_2,x_3))$& \\
  & \\
  \hline
\end{tabular}
\end{center}
In this case, the truth sub-tables of, formulas $\alpha_1=x_3\oplus(\neg x_1\otimes x_2)$, $\alpha_1=x_3\otimes(\neg x_1\otimes x_2)$ and $\alpha_1=\neg x_1\otimes(x_2\oplus x_3)$ are both $\lambda$-similar to $\psi_0(-x_1,x_2,x_3)$, where $\lambda=0.883$, since they differ in one position on 8 possible positions. This means that both formulas are 87.5\% accurate.

For a more complex configuration like $\alpha=\psi_0(-x_1,x_2,-x_3,x_4,-x_5)$, we can derive, using rule R, configurations:
\begin{center}
\tiny
\begin{tabular}{|c|c|}
  \hline
  & \\
  $\beta_1=\psi_0(-x_5,\psi_0(x_4,\psi_0(-x_3,\psi_0(x_2,-x_1))))$& $\beta_2=\psi_{-1}(x_4,\psi_{-1}(x_2,\psi_0(-x_5,\psi_0(-x_3,-x_1))))$\\
  $\beta_3=\psi_{-1}(x_4,\psi_{0}(-x_5,\psi_0(x_2,\psi_1(-x_3,-x_1))))$&
  $\beta_4=\psi_{-1}(x_4,\psi_{0}(x_2,\psi_0(-x_5,\psi_1(-x_3,-x_1))))$\\
  & \\
  \hline
\end{tabular}
\end{center}
Since these configurations are not equivalents, we concluded that $\alpha$ is un-representable. In this case we can see a change in the similarity level between $\alpha$ and each $\beta_i$ when the number of truth valued is changed:
\begin{center}
\tiny
\begin{tabular}{|l|l|}
  \hline
  & \\
  In the $2$-valued logic & $\alpha\sim_{0.8556}\beta_1$, $\alpha\sim_{0.9103}\beta_2$, $\alpha\sim_{0.5189}\beta_3$ and $\alpha\sim_{0.5880}\beta_4$ \\
  In the $3$-valued logic & $\alpha\sim_{0.8746}\beta_1$, $\alpha\sim_{0.9213}\beta_2$, $\alpha\sim_{0.4829}\beta_3$ and $\alpha\sim_{0.5483}\beta_4$ \\
  In the $4$-valued logic & $\alpha\sim_{0.8860}\beta_1$ , $\alpha\sim_{0.9268}\beta_2$, $\alpha\sim_{0.4667}\beta_3$ and $\alpha\sim_{0.5299}\beta_4$ \\
  In the $5$-valued logic & $\alpha\sim_{0.1120}\beta_1$, $\alpha\sim_{0.0710}\beta_2$, $\alpha\sim_{0.7810}\beta_3$ and $\alpha\sim_{0.6550}\beta_4$\\
  In the $10$-valued logic & $\alpha\sim_{0.0960}\beta_1$, $\alpha\sim_{0.0620}\beta_2$, $\alpha\sim_{0.8170}\beta_3$ and $\alpha\sim_{0.6950}\beta_4$\\
  & \\
  \hline
\end{tabular}
\end{center}

From observed similarity we selected $\beta_2$ as the best approximation to $\alpha$. Its quality, as an approximation, improves when we increase the logics number of truth values. Similarity increases with the increase in the number of evaluations.

For an un-representable configuration, $\alpha$, we can generate the finite set $S(\alpha)$, with representable networks similar to $\alpha$, using rule R. Given a $(n+1)$-valued logic, from that set of formulas we can select as an approximation to $\alpha$; the formula having the interpretation more similar to $\alpha$. This  identification of un-representable configuration, using representable approximations,  is used to transform networks with un-representable neurons into representable structures. The stress associated with this transformation characterizes the translation accuracy.

\subsection{Neural network crystallization}

Weights in CNNs assume the values -1 or  1. Naturally, every NN with weighs in $[-1,1]$ can be seen as an approximation to a CNNs. The process of identifying a NN with weighs in $[-1,1]$
as a {\L}NNs is called \emph{crystallization}, and essentially consists in rounding each neural weight $w_i$ to the nearest integer less than or equal to $w_i$, denoted by $\lfloor w_i\rfloor$.

 In this sense the crystallization process can be seen as a pruning
 on the network structure, where links between neurons with weights near 0 are removed and weights near -1 or 1 are consolidated. However this process is very crispy. We  need a smooth procedure to crystallize a network, in each learning iteration, to avoid the drastic reduction in learning performance.
 In each iteration we restricted the NN representation bias, making  the network representation bias converge to a structure similar to a CNN.
  For that, we defined by \emph{representation error} for a network $N$ with weights $w_1,\ldots,w_n$, as  $
 _{\Delta(N)=\sum^n_{i=1}(w_i-\lfloor w_i\rfloor).}
 $
 When $N$ is a CNNs we have $\Delta(N)=0$. Our \emph{smooth crystallization} process results from the iterating of function:
 \[
 _{\Upsilon_n(w)=sign(w).((\cos(1-abs(w)-\lfloor abs(w)\rfloor).\frac{\pi}{2})^n+\lfloor abs(w)\rfloor),}
 \]
where $sign(w)$ is the sign of $w$ and $abs(w)$ its absolute value. Denoting by $\Upsilon_n(N)$ the function having by input and output a NN, where the weights on the output network results of applying  $\Upsilon$ to all the input network weights and neurons biases. Each interactive application of $\Upsilon$ produce a networks progressively more similar to a CNNs. Since, for every network $N$ and $n>0$, $\Delta(N)\geq \Delta(\Upsilon_n(N))$, we have:

\begin{prop} Given a NNs $N$ with weights in the interval $[0,1]$. For every $n>0$ the function $\Upsilon_n(N)$ has, by fixed points, a CNNs.
\end{prop}

The convergence speed depends on parameter $n$. Increasing $n$ speeds up crystallization but reduces the network's plasticity to the training data.
For our applications, we selected $n=2$ based on the learning efficiency of a set of test formulas. Greater values for $n$ imposes stronger restrictions to learning. This procedure induces a quicker convergence to an admissible configuration of CNNs.

\section{\uppercase{Learning}}

Given a truth table on a $(n+1)$-valued {\L}logic, generated using a formula in the {\L}logic language, we will try to find its interpretation in the form of a {\L}NN, and from it, rediscover the original formula.

For that we trained a feed-forward NN using a truth table. Our methodology  trains progressively more complex networks until a crystallized network with good performance has been found. The  methodology is described in algorithm \ref{RevEng} that is used on the truth table reverse engineering task.

\begin{algorithm}
\caption{Reverse Engineering} \label{RevEng}
\begin{algorithmic}[1]
\tiny
\STATE Given a $(n+1)$-valued truth sub-table for a {\L}logic proposition
\STATE Define an inicial network complexity
\STATE Generate an inicial NN
\STATE Apply (the selected) Backpropagation algorithm using the data set
\IF{the generated network  have bad performance}
\STATE If need increase network complexity
\STATE Try a new network. Go to 3
\ENDIF
\STATE Do NN crystallization using the crisp process.
\IF{crystalized network have bad performance}
\STATE Try a new network. Go to 3
\ENDIF
\STATE Refine the crystalized network
\end{algorithmic}
\end{algorithm}

Given part of a truth table we try to find a {\L}NN that codifies the data. For this we generated NNs with a fixed number of hidden layers (our implementation uses three hidden layers). When the process detects bad learning performances, it aborts the training,   generating a new network with random heights. After a fixed number of tries, the network topology is changed. The number of tries for each topology depends on the number of network inputs. After trying to configure a set of networks for a given complexity with bad learning performance, the system tries to apply the selected back-propagation algorithm to a more complex set of networks. If the system finds a network codifying the data, the network is crystallized. When the error associated to this process increase,  the system returns to the learning phase and tries to configure a new network. When the process converges and the resulting network can be codified as a crisp {\L}NN the system prunes the network, for that we selected the ''Optimal Brain Surgeon'' algorithm proposed by G.J. Wolf, B. Hassibi and D.G. Stork in \cite{Hassibi93}.

\subsection{\uppercase{Training the neural network}}
Standard error back-propagation algorithm (EBP) is a gradient descent algorithm, in which the network weights are moved along the negative of the gradient of the performance function. EBP algorithm has been a significant improvement in NN research, but it has a weak convergence rate. Many efforts have been made to speed up the EBP algorithm. The Levenberg-Marquardt (LM) algorithm \cite{HaganMenhaj99} \cite{Andersen95} ensued from the development of EBP algorithm-dependent methods. It gives a good exchange between the speed of the Newton algorithm and the stability of the steepest descent method \cite{Battiti92}.

The basic EBP algorithm adjusts the weights in the steepest descent direction.
When training with the EBP method, an iteration of the algorithm defines the change of weights and has the form
$
_{w_{k+1}=w_k-\alpha G_k,}
$
where $G_k$ is the gradient of performance index $F$ on $w_k$, and $\alpha$ is the learning rate.

Note that the basic step of Newton's method can be derived from Taylor formula and is $
_{w_{k+1}=w_k-H_k^{-1}G_k,}
$
where $H_k$ is the Hessian matrix of the performance index at the current values of the weights.

Since Newton's method implicitly uses quadratic assumptions, the Hessian matrix dos not need be evaluated exactly. Rather, an approximation can be used, such as
$
_{H_k\approx J_k^TJ_k,}
$
where $J_k$ is the Jacobian matrix that contains first derivatives of the network errors with respect to the weights $w_k$.

The simple gradient descent and newtonian iteration are complementary in the advantages they provide. Levenberg proposed an algorithm based on this observation, whose update rule blends aforementioned algorithms and is given as
\[
_{w_{k+1}=w_k-[J_k^TJ_k+\mu I]^{-1}J_k^Te_k},
\]
where $J_k$ is the Jacobian matrix evaluated at $w_k$ and $\mu$ is the learning rate. This update rule is used as follows. If the error goes down following an update, it implies that our quadratic assumption on the function is working and we reduce $\mu$ (usually by a factor of 10) to reduce the influence of gradient descent. In this way, the performance function is always reduced at each iteration of the algorithm \cite{Megan96}. On the other hand, if the error goes up, we would like to follow the gradient more and so $\mu$ is increased by the same factor.

We can obtain some advantage out of the second derivative, by scaling each component of the gradient according to the curvature. This should result in larger movements along the direction where the gradient is smaller so the classic "error valley" problem does not occur any more. This crucial insight was provided by Marquardt. He replaced the identity matrix in the Levenberg update rule with the diagonal of Hessian matrix approximation resulting in the  LM update rule.
We changed the LM  algorithm by applying a soft crystallization step after the  LM update rule:
\[
_{w_{k+1}=\Upsilon_2(w_k-[J_k^TJ_k+\mu
.diag(J_k^TJ_k)]^{-1}J_k^Te_k})
\]
 This drastically improves the convergence to a CNN.

In our methodology network regularization is made using three different strategies:
\begin{enumerate}
  \item using soft crystallization, where knowledge dissemination is restricted on the network, information is concentrated on some weights;
  \item using crisp crystallization where only the heavier weights survive defines the network topology;
  \item pruning the resulting crystallized network.
\end{enumerate}
The last regularization technic avoids redundancies, in the sense that the same or redundant information can be codified at different locations. We minimized this by selecting weights to eliminate. For this task, we used "\emph{Optimal Brain Surgeon}" method, which uses the criterion of minimal increase in training error. It uses information from all second-order derivatives of the error function to perform network pruning.

\section{\uppercase{Applying reverse engineering on truth tables}}

Given a {\L}NN it can be translated in the form of a string base formula if every neuron is representable. Proposition \ref{conf classification} defines a tool to translate from the connectionist representation to a symbolic representation. It is remarkable that, when the truth table sample used in the learning was generated by a formula, the Reverse Engineering algorithm converges to a representable {\L}NN equivalent to the original formula, when evaluated on the cases used in the truth table sample.

When we generate a truth table in the $4$-valued  {\L}logic using formula
\[
\tiny
_{(x_4\otimes x_5\Rightarrow x_6)\otimes(x_1\otimes x_5\Rightarrow x_2)\otimes(x_1\otimes x_2\Rightarrow x_3)\otimes(x_6\Rightarrow x_4)}
\]
it has 4096 cases, the result of applying the algorithm is the 100\% accurate NN:
\[
\tiny
\begin{tabular}{lll}
  $\left[
    \begin{array}{cccccc}
      0 &  0 & 0 & -1 & 0 & 1 \\
      0 &  0 & 0 &  1 & 1 & -1\\
      1 &  1 & -1 & 0 &  0 & 0\\
      -1 &  1 & 0 & 0 &  -1 & 0\\
    \end{array}
  \right]
  $ &$ \left[
       \begin{array}{c}
         0 \\
         -1 \\
         -1 \\
         2 \\
       \end{array}
     \right]
    $& $\begin{array}{l}
         \neg x_4\otimes x_6 \\
         x_4\otimes x_5 \otimes \neg x_6  \\
         x_1\otimes x_2 \otimes \neg x_3 \\
         \neg x_1\oplus x_2 \oplus \neg x_5
       \end{array}$\\
  $\left[
    \begin{array}{cccc}
      -1 &  -1 & -1 & 1\\
    \end{array}
  \right]
   $&$ \left[
       \begin{array}{c}
         0 \\
       \end{array}
     \right]
    $& $\begin{array}{l}
         \neg i_1 \otimes \neg i_2 \otimes \neg i_3 \otimes i_4 \\
       \end{array}$\\
  $\left[
       \begin{array}{c}
          1\\
       \end{array}
     \right] $& $\left[
       \begin{array}{c}
         0 \\
       \end{array}
     \right]$ & $j_1$\\
\end{tabular}
\]
Using local interpretation we may reconstruct the formula:
\begin{center}
\tiny
$
j_1 = \neg i_1 \otimes \neg i_2 \otimes \neg i_3 \otimes i_4 =
$
$
\neg (\neg x_4\otimes x_6) \otimes \neg (x_4\otimes x_5 \otimes \neg x_6) \otimes \neg (x_1\otimes x_2 \otimes \neg x_3) \otimes (\neg x_1\oplus x_2 \oplus \neg x_5)=
$
$
= (x_4\oplus \neg x_6) \otimes (\neg x_4\oplus \neg x_5 \oplus  x_6) \otimes  (\neg x_1\oplus \neg x_2 \oplus x_3) \otimes (\neg x_1\oplus x_2 \oplus \neg x_5)=
$
$
= (x_6 \Rightarrow x_4) \otimes (x_4\otimes x_5 \Rightarrow x_6) \otimes (x_1 \otimes x_2 \Rightarrow x_3) \otimes (x_1 \otimes x_5 \Rightarrow x2)
$
\end{center}

The number of layers, used on our implementation, imposes structural restrictions formula reconstruction. A truth table generated by
$(((i_1\otimes i_2) \oplus (i_2 \otimes i_3))\otimes ((i_3\otimes i_4)\oplus (i_4\otimes i_5))) \oplus (i_5\otimes i_6)$
requires at least 4 hidden layers, to be reconstructed; this is the number of levels required by the associated parsing tree.
\begin{table}
\begin{center}
\tiny
\begin{tabular}{|r||r|r|}
  \hline
  formula & mean & stdev \\
  \hline
  $i_1\otimes i_3\Rightarrow i_6$ & 7.68 & 6.27 \\
  $i_4\Rightarrow i_6\otimes i_6\Rightarrow i_2$ & 25.53 & 11.14 \\
  $((i_1\Rightarrow i_4)\oplus(i_6\Rightarrow i_2))\otimes(i_6\Rightarrow i_1)$ & 43.27 & 14.25 \\
  $(i_4\otimes i_5\Rightarrow i_6)\otimes(i_1\otimes i_5\Rightarrow i_2)$ & 51.67 & 483.85 \\
  $((i_4\otimes i_5\Rightarrow i_6)\oplus(i_1\otimes i_5\Rightarrow i_2))\otimes(i_1\otimes i_3\Rightarrow i_2)$ & 268.31 & 190.99 \\
  $((i_4\otimes i_5\Rightarrow i_6)\oplus(i_1\otimes i_5\Rightarrow i_2))\otimes(i_1\otimes i_3\Rightarrow i_2)\otimes(i_6\Rightarrow i_4)$ & 410.47 & 235.52 \\
  \hline
\end{tabular}
\end{center}
\caption{Reverse engineering test formulas.}\label{revtest}
\end{table}

Table \ref{revtest} presents the mean CPU times need to find a configuration with a mean square error of less than 0.002. The mean time is computed using  6 trials on a 5-valued truth  {\L}logic for each formula. We implemented the algorithm using the MatLab neural network package and executed it in an AMD Athlon 64 X2 Dual-Core Processor TK-53 at 1.70 GHz on a Windows Vista system with 959MB of  memory.
In table \ref{revtest} the last two formula was approximated, since its complexity exceeds the structures modifiable on a NNs with three hidden layers. For the others formules the extraction process made equivalent reconstructions.

\section{\uppercase{Applying the process on real data}}

The described extraction process, when applied to real data, expresses the information using CNNs. This naturally means that the process searches for simple and understandable models for the data, able to be codify directly or approximated using {\L}logic first-order language. The process gives preference to the simplest models and subject them to a strong pruning criteria. With this strategy we avoid overfetting and the problems associated with the algorithm complexity.

\subsubsection{Mushrooms}
\emph{Mushroom} is a data set available in the \emph{UCI Machine Learning Repository}. This data set includes descriptions of hypothetical samples corresponding to 23 species of gilled mushrooms in the Agaricus and Lepiota Family. Each species is identified as definitely edible, definitely poisonous, or of unknown edibility and not recommended. This latter class was combined with the poisonous one. The Guide clearly states that there is no simple rule for determining the edibility of a mushroom. However, we will try to find one using the data set as a truth table.

The data set has 8124 instances defined using 22 nominally valued attributes presented in the table below. It has missing attribute values, 2480, all for attribute \#11. 4208 instances (51.8\%) are classified as edible and 3916 (48.2\%) are classified as poisonous.

We used an unsupervised filter that converted all nominal attributes into binary
numeric attributes. An attribute with $k$ values was transformed into $k$
binary attributes. This produced a data set containing 111 binary attributes.

After the binarization we used the described method to select relevant attributes for mushroom classification by fixing a weak stoping criterion. As a result, the method produced a model, with 100\% accuracy, depending  on 23 binary attributes defined by values of:
odor,gill.size,stalk.surface.above.ring, ring.type, spore.print.color.

We used the values assumed by these attributes to produce a new data set. After 3 tries we selected the model less complex:
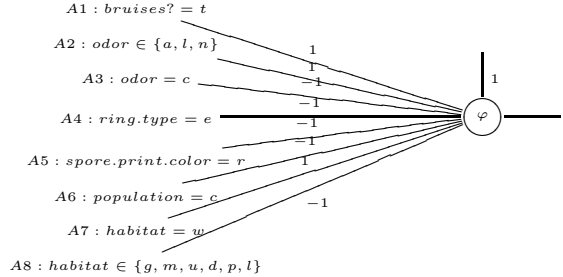
\[
\tiny
\xymatrix @R=3pt @C=20pt
     {  A1: bruises? = t \ar@{-}[dddrrr]^{1}&&&  & \\
        A2: odor\in\{a,l,n\}\ar@{-}[ddrrr]^{1} &&& \ar@{-}[dd]^{1} & \\
        A3: odor=c \ar@{-}[drrr]^{-1}&&&  & \\
        A4: ring.type = e \ar@{-}[rrr]^{-1}   &&& *++[o][F-]{\varphi} \ar@{-}[r]& \\
        A5: spore.print.color = r \ar@{-}[urrr]^{-1}&&&  & \\
        A6: population = c \ar@{-}[uurrr]^{-1}&&&  & \\
        A7: habitat = w\ar@{-}[uuurrr]^{1} &&&  & \\
        A8: habitat \in \{g,m,u,d,p,l\} \ar@{-}[uuuurrr]_{-1}&&&  &  \
     }
\]

This model has an accuracy of 100\%. From it, and since attribute values in A2 and A3, as well as the values in A7 and A8 are
auto-exclusive, we used propositions A1, A2, A3, A4, A5, A6 and A7 to define a new data set. This new data set was
enriched with new negative cases by introducing, for each original case, a new one where the
truth value of each attribute was multiplied by 0.5. For instance, the ''eatable'' mushroom case:
\begin{center}
\tiny
 (A1=0, A2=1, A3=0, A4=0, A5=0, A6=0, A7=0,A8=1,A9=0)
\end{center}
was used on the definition of a new ''poison'' case
\begin{center}
\tiny
 (A1=0, A2=0.5, A3=0, A4=0, A5=0, A6=0, A7=0,A8=0.5,A9=0)
\end{center}
This resulted in a convergence speedup and reduced the occurrence of un-representable configurations.
\begin{table}
\begin{center}
\tiny
\begin{tabular}{|l|l|l|}
  \hline
  N. & Attribute &Values \\
  \hline
  0& classes & edible=e, poisonous=p \\
  1& cap.shape & bell=b,conical=c,convex=x,flat=f,knobbed=k, sunken=s\\
  2& cap.surface & fibrous=f,grooves=g,scaly=y,smooth=s \\
  3& cap.color & brown=n,buff=b,cinnamon=c,gray=g,green=r, pink=p,purple=u,red=e,white=w,yellow=y\\
  4& bruises? & bruises=t,no=f \\
  5& odor & almond=a,anise=l,creosote=c,fishy=y,foul=f,musty=m,none=n,pungent=p,spicy=s\\
  6& gill.attachment & attached=a,descending=d,free=f,notched=n \\
  7& gill.spacing & close=c,crowded=w,distant=d \\
  8& gill.size & broad=b,narrow=n \\
  9& gill.color & black=k,brown=n,buff=b,chocolate=h,gray=g,green=r,orange=o,pink=p,purple=u,red=e,\\
  &&white=w,yellow=y \\
  10& stalk.shape & enlarging=e,tapering=t \\
  11& stalk.root & bulbous=b,club=c,cup=u,equal=e,rhizomorphs=z,rooted=r,missing=?\\
  12& stalk.surface.above.ring & ibrous=f,scaly=y,silky=k,smooth=s \\
  13& stalk.surface.below.ring & ibrous=f,scaly=y,silky=k,smooth=s \\
  14& stalk.color.above.ring & brown=n,buff=b,cinnamon=c,gray=g,orange=o, pink=p,red=e,white=w,yellow=y\\
  15& stalk.color.below.ring & brown=n,buff=b,cinnamon=c,gray=g,orange=o, pink=p,red=e,white=w,yellow=y\\
  16& veil.type & partial=p,universal=u \\
  17& veil.color & brown=n,orange=o,white=w,yellow=y \\
  18& ring.number & none=n,one=o,two=t \\
  19& ring.type & cobwebby=c,evanescent=e,flaring=f,large=l,none=n,pendant=p,sheathing=s,zone=z\\
  20& spore.print.color & black=k,brown=n,buff=b,chocolate=h,green=r,orange=o,purple=u,white=w,yellow=y\\
  21& population & abundant=a,clustered=c,numerous=n,scattered=s, several=v,solitary=y\\
  22& habitat & grasses=g,leaves=l,meadows=m,paths=p,urban=u,waste=w,woods=d\\
  \hline
\end{tabular}
\end{center}
\caption{\emph{Mushroom} data set attribute Information.}
\end{table}

When we applied our "reverse engineering" algorithm to the enriched data set, having as stopping criterion the mean square error ($mse$) less than $0.003$, the method produced the model:
\[
\tiny
\begin{tabular}{lll}
  $\left[
    \begin{array}{ccccccc}
      0 &  1 & 0 & 0 & -1 & 0 & 1 \\
      0 &  1 & 0 & 1 &  0 & 0 & -1 \\
    \end{array}
  \right]
  $ &$ \left[
       \begin{array}{c}
         -1 \\
         -1 \\
       \end{array}
     \right]
    $& $\begin{array}{l}
         A2\otimes \neg A5 \otimes A7 \\
         A2\otimes A4\otimes \neg A7  \\
       \end{array}$\\
  $\left[
    \begin{array}{cc}
      1 & 1 \\
    \end{array}
  \right]
   $&$ \left[
       \begin{array}{c}
         0 \\
       \end{array}
     \right]
    $& $i_1\oplus i_2$\\
  $\left[
       \begin{array}{c}
         1 \\
       \end{array}
     \right] $& $\left[
       \begin{array}{c}
         0 \\
       \end{array}
     \right]$ \\
\end{tabular}
\]
This model codifies the proposition
$
_{(A2\otimes \neg A5 \otimes A7)\oplus(A2\otimes A4\otimes \neg A8)}
$
and misses the classification of 48 cases. It has 98.9\% accuracy.

More precise model can be produced, by restricting the stopping criteria. However, this in general, produces more complex propositions and is more difficult to understand. For instance with a stopping criterion $mse<0.002$ the systems generated the below model. It misses 32 cases, has an accuracy of 99.2\%, and it is easy to convert in a proposition.
\[
\tiny
\begin{tabular}{lll}
  $\left[
    \begin{array}{ccccccc}
      0 &  0 & 0 & -1 & 0 & 0 & 1 \\
      1 &  1 & 0 & -1 &  0 & 0 & 0 \\
      0 &  0 & 0 & 0 &  0 & 0 & 1 \\
      0 &  1 & 0 & 0 &  -1 & -1 & 1 \\
    \end{array}
  \right]
  $ &$ \left[
       \begin{array}{c}
         1 \\
         -1 \\
         0 \\
         -1\\
       \end{array}
     \right]
    $& $\begin{array}{l}
         \neg A4 \oplus A7 \\
         A1\otimes A2\otimes \neg A4  \\
         A7\\
         A2\otimes \neg A5\otimes \neg A6\otimes A7\\
       \end{array}$\\
  $\left[
    \begin{array}{cccc}
      -1 &  0 & 1 &  0 \\
       1 & -1 & 0 & -1 \\
    \end{array}
  \right]
   $&$ \left[
       \begin{array}{c}
         1 \\
         0 \\
       \end{array}
     \right]
    $& $\begin{array}{l}
         \neg i_1\oplus i_3 \\
         i_1\otimes \neg i_2 \otimes \neg i_4 \\
       \end{array}$\\
  $\left[
       \begin{array}{cc}
         1 & -1 \\
       \end{array}
     \right] $& $\left[
       \begin{array}{c}
         0 \\
       \end{array}
     \right]$ & $j_1\otimes\neg j_2$\\
\end{tabular}
\]
This NN can be used to interprete formula:
\begin{center}
\tiny
$j_1\otimes\neg j_2 =((A4 \otimes \neg A7)\oplus A7)\otimes (( A4 \otimes \neg A7)\oplus (A1\otimes A2\otimes \neg A4) \oplus (A2\otimes \neg A5\otimes \neg A6\otimes A7))
$
\end{center}

Some times the algorithm converged to un-representable configurations like the one presented below, with 100\% accuracy. The frequency of this type of configurations increases with the increase of required accuracy.
\[
\tiny
\begin{tabular}{lll}
  $\left[
    \begin{array}{ccccccc}
      -1 &  1 & -1 & 1 & 0 & -1 & 0 \\
      0 &  0 & 0 & 1 &  1 & 0 & -1 \\
      1 &  1 & 0 & 0 &  0 & 0 & -1 \\
    \end{array}
  \right]
  $ &$ \left[
       \begin{array}{c}
         0 \\
         1 \\
         0 \\
       \end{array}
     \right]
    $& $\begin{array}{l}
         i_1\text{ un-representable} \\
         A4\otimes A5\otimes \neg A6  \\
         i_3\text{ un-representable} \\
       \end{array}$\\
  $\left[
    \begin{array}{ccc}
      1 &  -1 & 1 \\
    \end{array}
  \right]
   $&$ \left[
       \begin{array}{c}
         0 \\
       \end{array}
     \right]
    $& $\begin{array}{l}
         j_1 \text{un-representable} \\
       \end{array}$\\
  $\left[
       \begin{array}{c}
         1 \\
       \end{array}
     \right] $& $\left[
       \begin{array}{c}
         0 \\
       \end{array}
     \right]$ & \\
\end{tabular}
\]
Using rule R and selecting the best approximation in data set to each un-representable formula, evaluated in the data set, we have:
\begin{center}
\tiny
\begin{tabular}{|c|c|c|}
  \hline
  & &\\
  $i_1\sim_{0.9297}((\neg A1\otimes A4) \oplus A2)\otimes \neg A3 \otimes \neg A6$ & $i_3\sim_{1.0}(A1\oplus\neg A7)\otimes A2$ &
  $j_1\sim_{0.9951}(i_1\otimes\neg i_2)\oplus i_3$ \\
  & &\\
  \hline
\end{tabular}
\end{center}
The extracted formula
\begin{center}
\tiny
$\alpha=(((((\neg A1\otimes A4) \oplus A2)\otimes \neg A3 \otimes \neg A6)\otimes\neg (A4\otimes A5\otimes \neg A6))\oplus ((A1\oplus\neg A7)\otimes A2)$
\end{center}
is $\lambda$-similar, with $\lambda=0.9951$ to the original NN. Formula $\alpha$ misses the classification for 40 cases. Note that the symbolic model is stable, the bad performance of $i_1$ representation do not affect the model.

\section{Conclusions}
This methodology to codify and extract symbolic knowledge from a NN is very simple and efficient for the extraction of comprehensible rules from medium-sized data sets. It is, moreover, very sensible to attribute relevance.

In the theoretical point of view it is particularly interesting that restricting the values assumed by neurons weights restrict the information propagation in the network, thus allowing the emergence of patterns in the neuronal network structure. For the case of linear neuronal networks, having by activation function the identity truncate to 0 and 1, these structures are characterized by the occurrence of patterns in neuron configuration  directly presentable as formulas in {\L}logic.

\bibliographystyle{splncs}
{\tiny
\bibliography{multibib}}

\begin{thebibliography}{10}

\bibitem{Castro98}
Castro, J., Trillas, E.:
\newblock The logic of neural networks.
\newblock Mathware and Soft Computing, vol. 5, (1998)23-27. (1998)

\bibitem{Hitzler04}
Hitzler, P., H\"{o}lldobler, S., Seda, A.:
\newblock Logic programs and connectionist networks.
\newblock Journal of Applied Logic, 2, (2004)245-272. (2004)

\bibitem{Holldobler00}
H\"{o}lldobler, S.:
\newblock Challenge problems for the integration of logic and connectionist
  systems.
\newblock in: F. Bry, U.Geske and D. Seipel, editors, Proceedings 14. Workshop
  Logische Programmierung, GMD Report 90, (2000)161-171. (2000)

\bibitem{Fu93}
Fu, L.:
\newblock Knowledge-based connectionism from revising domain theories.
\newblock IEEE Trans. Syst. Man. Cybern, Vol. 23 ,(1993)173-182. (1993)

\bibitem{Towell94}
Towell, G., Shavlik, J.:
\newblock Knowledge-based artificial neural networks.
\newblock Artif. Intell., Vol. 70 ,(1994)119-165. (1994)

\bibitem{Gallan88}
Gallant, S.:
\newblock Connectionist expert systems.
\newblock Commun. ACM, Vol. 31 ,(1988)152-169. (1988)

\bibitem{Gallan94}
Gallant, S.:
\newblock Neural Network Learning and Expert Systems.
\newblock Cambridge, MA, MIT Press (1994)

\bibitem{Towell93}
Towell, G., Shavlik, J.:
\newblock Extracting refined rules from knowledge-based neural networks.
\newblock Mach. Learn., Vol. 13 ,(1993)71-101. (1993)

\bibitem{HaganMenhaj99}
Hagan, M., Menhaj, M.:
\newblock Training feed-forward networks with marquardt algorithm.
\newblock IEEE Transaction on Neural Networks, vol. 5 no. 6, (1999)989-993.
  (1999)

\bibitem{Hassibi93}
Hassibi, B., Stork, D., Wolf, G.:
\newblock Optimal brain surgeon and general network pruning.
\newblock IEEE International Conference on Neural Network, vol. 4 no. 5,
  (2003)740-747. (1993)

\bibitem{Hajek95}
H\'{a}jek, P.:
\newblock Fuzzy logic from the logical point of view.
\newblock In Proceedings SOFSEM'95, LNCS, Springer-Verlag, 1995. (1995)

\bibitem{Amato02}
Amato, P., Nola, A., Gerla, B.:
\newblock Neural networks and rational {\l}ukasiewicz logic.
\newblock IEEE Transaction on Neural Networks, vol. 5 no. 6, (2002)506-510.
  (2002)

\bibitem{Fiadeiro97}
Fiadeiro, J., Lopes, A.:
\newblock Semantics of architectural connectors.
\newblock TAPSOFT'97 LNCS, v.1214, p.505-519, Springer-Verlag, 1997. (1997)

\bibitem{Andersen95}
Andersen, T., Wilamowski, B.:
\newblock A modified regression algorithm for fast one layer neural network
  training.
\newblock World Congress of Neural Networks, Washington DC, USA, Vol. 1 no. 4,
  CA, (1995)687-690. (1995)

\bibitem{Battiti92}
Battiti, R.:
\newblock Frist- and second-order methods for learning between steepest descent
  and newton's method.
\newblock Neural Computation, Vol. 4 no. 2, (1992)141-166. (1992)

\bibitem{Megan96}
Hagan, M., Demuth, H., Beal, M.:
\newblock Neural Network Design.
\newblock PWS Publishing Company, Boston (1996)

\end{thebibliography}
\end{document}